\documentclass[10pt,twocolumn,letterpaper]{article}

\usepackage{cvpr}
\usepackage{times}
\usepackage{epsfig}
\usepackage{graphicx}
\usepackage{amsmath}
\usepackage{amssymb}

\usepackage{bbm}
\usepackage{array}
\usepackage{color}
\usepackage{multirow}
\usepackage[tight,normalsize,sf,SF]{subfigure}
\usepackage[pagebackref=true,breaklinks=true,letterpaper=true,colorlinks,bookmarks=false]{hyperref}

\cvprfinalcopy % *** Uncomment this line for the final submission

\makeatletter\renewcommand\paragraph{\@startsection{paragraph}{4}{\z@}
  {.5em \@plus1ex \@minus.2ex}{-.5em}{\normalfont\normalsize\bfseries}}\makeatother
  
\def\cvprPaperID{1673} % *** Enter the CVPR Paper ID here

\ifcvprfinal\fi
\begin{document}

%%%%%%%%% TITLE
\title{Improving Transferability of Adversarial Examples with Input Diversity}

\author{
Cihang Xie\textsuperscript{1} \qquad
Zhishuai Zhang\textsuperscript{1} \qquad
Yuyin Zhou\textsuperscript{1} \qquad
Song Bai\textsuperscript{2} \\
Jianyu Wang\textsuperscript{3} \qquad
Zhou Ren\textsuperscript{4} \qquad
Alan Yuille\textsuperscript{1} \vspace{.5em}\\
\textsuperscript{1}Johns Hopkins University \quad  \textsuperscript{2}University of Oxford \quad \textsuperscript{3}Baidu Research \quad \textsuperscript{4}Wormpex AI Research}

\maketitle

\thispagestyle{empty}

\begin{abstract}
Though CNNs have achieved the state-of-the-art performance on various vision tasks, they are vulnerable to adversarial examples --- crafted by adding human-imperceptible perturbations to clean images. However, most of the existing adversarial attacks only achieve relatively low success rates under the challenging black-box setting, where the attackers have no knowledge of the model structure and parameters. To this end, we propose to improve the transferability of adversarial examples by creating diverse input patterns. Instead of only using the original images to generate adversarial examples, our method applies random transformations to the input images at each iteration. Extensive experiments on ImageNet show that the proposed attack method can generate adversarial examples that transfer much better to different networks than existing baselines. 
By evaluating our method against top defense solutions and official baselines from NIPS $2017$ adversarial competition, the enhanced attack reaches an average success rate of $73.0\%$, which outperforms the top-$1$ attack submission in the NIPS competition by a large margin of $6.6\%$. We hope that our proposed attack strategy can serve as a strong benchmark baseline for evaluating the robustness of networks to adversaries and the effectiveness of different defense methods in the future. Code is available at \url{https://github.com/cihangxie/DI-2-FGSM}.
\end{abstract}

\section{Introduction}
Recent success of Convolutional Neural Networks (CNNs) leads to a dramatic performance improvement on various vision tasks, including image classification~\cite{krizhevsky2012imagenet,simonyan2015very,he2016identity}, object detection~\cite{Girshick_2015_Fast,Ren_2015_Faster,zhang2017single} and semantic segmentation~\cite{Long_2015_Fully,Chen_2016_DeepLab}. However, CNNs are extremely vulnerable to small perturbations to the input images,~\ie,~human-imperceptible additive perturbations can result in failure predictions of CNNs. These intentionally crafted images are known as adversarial examples~\cite{szegedy2013intriguing}. Learning how to generate adversarial examples can help us investigate the robustness of different models~\cite{arnab2017robustness} and understand the insufficiency of current training algorithms~\cite{goodfellow2014explaining,kurakin2016scale,tramer2017ensemble}.  

\begin{figure}[tb] 
\centering
\includegraphics[width=\linewidth]{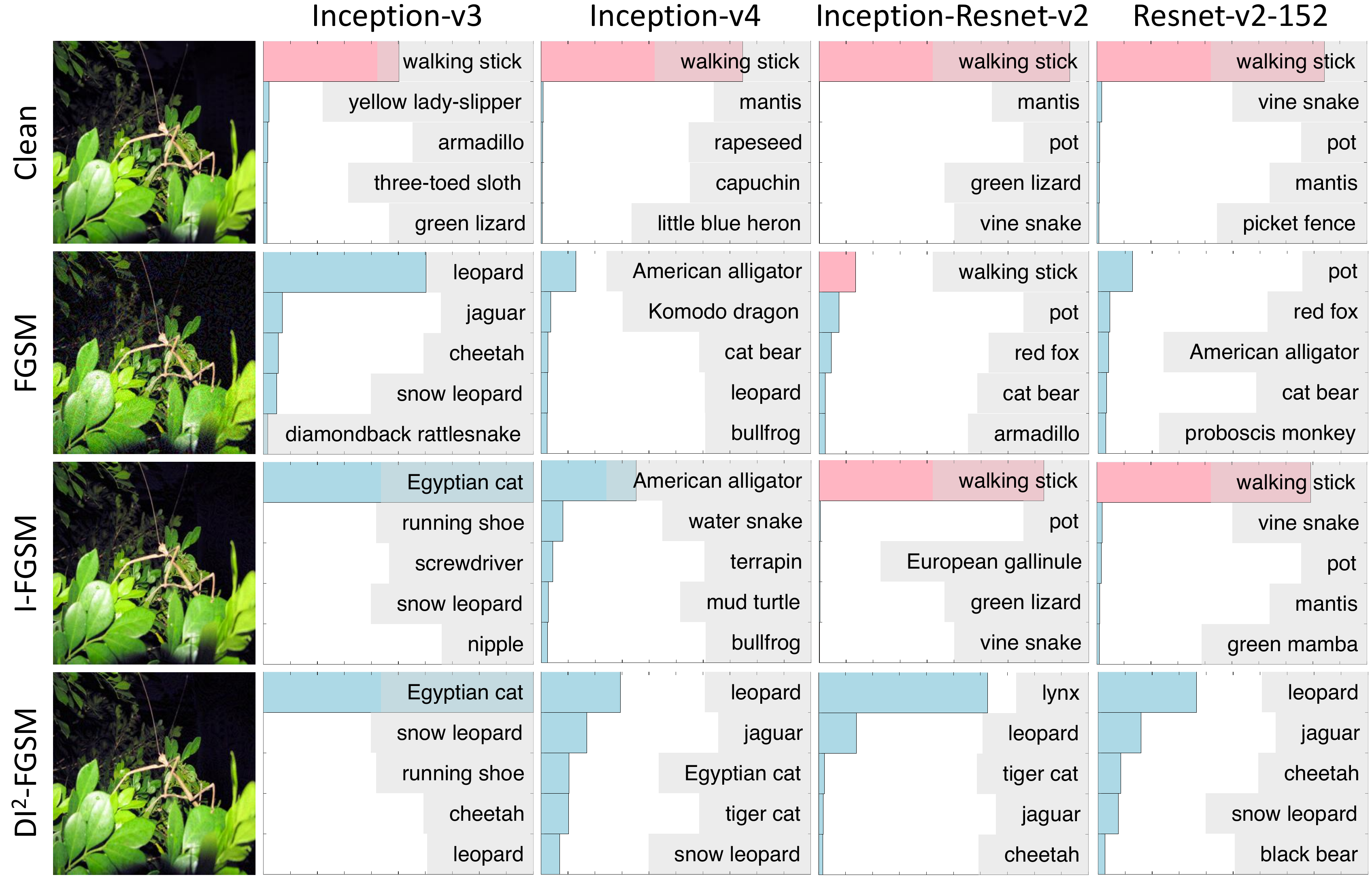}
\caption{\textbf{The comparison of success rates using three different attacks}. The ground-truth ``walking stick" is marked as pink in the top-$5$ confidence distribution plots. The adversarial examples are crafted on Inception-v3 with the maximum perturbation $\epsilon=15$. From the first row to the the third row, we plot the top-$5$ confidence distributions of clean images, FGSM and I-FGSM, respectively. The fourth row shows the result of the proposed Diverse Inputs Iterative Fast Gradient Sign Method (DI\textsuperscript{2}-FGSM), which attacks the white-box model and all black-box models successfully.}
\label{fig:illustration}
\vspace{-1em}
\end{figure}

Several methods~\cite{goodfellow2014explaining,szegedy2013intriguing,kurakin2016adversarial} have been proposed recently to find adversarial examples. In general, these attacks can be categorized into two types according to the number of steps of gradient computation,~\ie,~single-step attacks~\cite{goodfellow2014explaining} and iterative attacks~\cite{szegedy2013intriguing,kurakin2016adversarial}. Generally, iterative attacks can achieve higher success rates than single-step attacks in the white-box setting, where the attackers have a perfect knowledge of the network structure and weights. However, if these adversarial examples are tested on a different network (either in terms of network structure, weights or both),~\ie,~the black-box setting, single-step attacks perform better. This trade-off is due to the fact that iterative attacks tend to overfit the specific network parameters (\ie, have high white-box success rates) and thus making generated adversarial examples rarely transfer to other networks (\ie, have low black-box success rates), while single-step attacks usually underfit to the network parameters (\ie, have low white-box success rates) thus producing adversarial examples with slightly better transferability. Observing the phenomenon, one interesting question is whether we can generate adversarial examples with high success rates under both white-box and black-box settings. 

In this work, we propose to improve the transferability of adversarial examples by creating diverse input patterns. Our work is inspired by the data augmentation~\cite{krizhevsky2012imagenet,simonyan2015very,he2016identity} strategy, which has been proven effective to prevent networks from overfitting by applying a set of label-preserving transformations (\eg, resizing, cropping and rotating) to training images. 
Meanwhile, \cite{xie2017mitigating,guo2017countering} showed that image transformations can defend against adversarial examples under certain situations, which indicates adversarial examples cannot generalize well under different transformations. These transformed adversarial examples are known as hard examples~\cite{shrivastava2016training,simo2015discriminative} for attackers, which can then be served as good samples to produce more transferable adversarial examples.

We incorporate the proposed input diversity strategy with iterative attacks,~\eg,~I-FGSM~\cite{kurakin2016scale} and MI-FGSM~\cite{dong2017boosting}. At each iteration, unlike the traditional methods which maximize the loss function directly w.r.t. the original inputs, we apply random and differentiable transformations (\eg,~random resizing, random padding) to the input images with probability $p$ and maximize the loss function w.r.t. these transformed inputs. 
Note that these randomized operations were previously used to defend against adversarial examples~\cite{xie2017mitigating}, while here we incorporate them into the attack process to create hard and diverse input patterns. Fig.~\ref{fig:illustration} shows an adversarial example generated by our method and compares the success rates to other attack methods under both white-box and black-box settings.

We test the proposed input diversity on several network under both white-box and black-box settings, and single-model and multi-model settings. Compared with traditional iterative attacks, the results on ImageNet (see Sec.~\ref{sec: single model}) show that our method gets significantly higher success rates for black-box models and maintains similar success rates for white-box models. By evaluating our attack method w.r.t. the top defense solutions and official baselines from NIPS $2017$ adversarial competition~\cite{kurakin2018adversarial}, this enhanced attack reaches an average success rate of $73.0\%$, which outperforms the top-$1$ attack submission in the NIPS competition by a large margin of $6.6\%$. We hope that our proposed attack strategy can serve as a benchmark for evaluating the robustness of networks to adversaries and the effectiveness of different defense methods in future.

\section{Related Work}
\subsection{Generating Adversarial Examples}
Traditional machine learning algorithms are known to be vulnerable to adversarial examples~\cite{dalvi2004adversarial,huang2011adversarial,biggio2013evasion}. Recently, Szegedy~\etal~\cite{szegedy2013intriguing} pointed out that CNNs are also fragile to adversarial examples, and proposed a box-constrained L-BFGS method to find adversarial examples reliably. Due to the expensive computation in \cite{szegedy2013intriguing}, Goodfellow~\etal~\cite{goodfellow2014explaining} proposed the fast gradient sign method to generate adversarial examples efficiently by performing a single gradient step. This method was extended by Kurakin \etal \cite{kurakin2016adversarial} to an iterative version, and showed that the generated adversarial examples can exist in the physical world. Dong~\etal~\cite{dong2017boosting} proposed a broad class of momentum-based iterative algorithms to boost the transferability of adversarial examples. The transferability can also be improved by attacking an ensemble of networks simultaneously~\cite{liu2016delving}. Besides image classification, adversarial examples also exist in object detection~\cite{xie2017adversarial}, semantic segmentation~\cite{xie2017adversarial,cisse2017houdini}, speech recognition~\cite{cisse2017houdini}, deep reinforcement learning~\cite{lin2017tactics}, etc.. Unlike adversarial examples which can be recognized by human, Nguyen~\etal~\cite{nguyen2015deep} generated fooling images that are different from natural images and difficult for human to recognize, but CNNs classify these images with high confidences.

Our proposed input diversity is also related to EOT \cite{athalye2018synthesizing}. These two works differ in several aspects: (1) we mainly focus on the challenging black-box setting while \cite{athalye2018synthesizing} focuses on the white-box setting; (2) our work aims at alleviating overfitting in adversarial attacks, while \cite{athalye2018synthesizing} aims at making adversarial examples robust to transformations, without any discussion of overfitting; and (3) we do not apply expectation step in each attack iteration, while ``expectation'' is the core idea in \cite{athalye2018synthesizing}. 

\subsection{Defending Against Adversarial Examples}
Conversely, many methods have been proposed recently to defend against adversarial examples. \cite{goodfellow2014explaining,kurakin2016scale} proposed to inject adversarial examples into the training data to increase the network robustness. Tram{\`e}r~\etal~\cite{tramer2017ensemble} pointed out that 
such adversarially trained models still remain vulnerable to adversarial examples, and proposed ensemble adversarial training, which augments training data with perturbations transferred from other models, in order to improve the network robustness further. \cite{xie2017mitigating,guo2017countering} utilized randomized image transformations to inputs at inference time to mitigate adversarial effects. Dhillon~\etal~\cite{s2018stochastic} pruned a random subset of activations according to their magnitude to  enhance network robustness. Prakash~\etal~\cite{prakash2018deflecting} proposed a framework which combines pixel deflection with soft wavelet denoising to defend against adversarial examples. \cite{meng2017magnet,song2017pixeldefend,samangouei2018defensegan} leveraged generative models to purify adversarial images by moving them back towards the distribution of clean images. 

\section{Methodology}
Let $X$ denote an image, and $y^{\text{true}}$ denote the corresponding ground-truth label. We use
$\theta$ to denote the network parameters, and $L(X, y^{\text{true}}; \theta)$ to denote the loss. To generate the adversarial example, the goal is to maximize the loss $L(X + r, y^{\text{true}}; \theta)$ for the image $X$, under the constraint that the generated adversarial example $X^{\text{adv}}=X+r$ should look visually similar to the original image $X$ and the corresponding predicted label $y^{\text{adv}}\neq y^{\text{true}}$. In this work, we use $l_{\infty}$-norm to measure the perceptibility of adversarial perturbations,~\ie,~$||r||_{\infty}\leq\epsilon$. The loss function is defined as
\begin{equation}
L(X,y^{true};\theta)=-\mathbbm{1}_{y^{true}}\cdot\log\left(\text{softmax}(l(X; \theta))\right),
\end{equation}
where $\mathbbm{1}_{y^{true}}$ is the one-hot encoding of the ground-truth $y^{true}$ and $l(X; \theta)$ is the logits output. Note that all the baseline attacks have been implemented in the cleverhans library~\cite{papernot2016cleverhans}, which can be used directly for our experiments.

\subsection{Family of Fast Gradient Sign Methods}
In this section, we give an overview of the family of fast gradient sign methods.
\paragraph{Fast Gradient Sign Method (FGSM).} FGSM~\cite{goodfellow2014explaining} is the first member in this attack family, which finds the adversarial perturbations in the direction of the loss gradient $\nabla_{X} L(X, y^{\text{true}}; \theta)$. The update equation is 
\begin{equation}
X^{\text{adv}} = X + \epsilon\cdot\text{sign}( \nabla_{X} L(X, y^{\text{true}}; \theta)).
\end{equation}

\paragraph{Iterative Fast Gradient Sign Method (I-FGSM).} Kurakin~\etal~\cite{kurakin2016scale} extended FGSM to an iterative version, which can be expressed as
\begin{align} \label{eq:IFGSM update}
\small
&X_{0}^{\text{adv}} = X   \\ 
&X_{n+1}^{\text{adv}} = \text{Clip}_{X}^{\epsilon} \left \{X_{n}^{adv} + \alpha\cdot\text{sign}(\nabla_{X} L(X_{n}^{adv}, y^{\text{true}}; \theta)) \right \}, \nonumber 
\end{align}
where $\text{Clip}_{X}^{\epsilon}$ indicates the resulting image are clipped within the $\epsilon$-ball of the original image $X$, $n$ is the iteration number and $\alpha$ is the step size.

\paragraph{Momentum Iterative Fast Gradient Sign Method (MI-FGSM).} MI-FGSM~\cite{dong2017boosting} proposed to integrate the momentum term into the attack process to stabilize update directions and escape from poor local maxima. The updating procedure is similar to I-FGSM, with the replacement of Eq.~\eqref{eq:IFGSM update} by:
\begin{equation} \label{eq: momentum term}
\begin{split}
&g_{n+1} = \mu \cdot g_n + \frac{\nabla_{X} L(X_{n}^{adv}, y^{\text{true}}; \theta)}{||\nabla_{X} L(X_{n}^{adv}, y^{\text{true}}; \theta)||_1} \\
&X_{n+1}^{\text{adv}} = \text{Clip}_{X}^{\epsilon} \left \{X_{n}^{adv} + \alpha \cdot \text{sign}(g_{n+1}) \right \},   
\end{split}
\end{equation}
where $\mu$ is the decay factor of the momentum term and $g_n$ is the accumulated gradient at iteration $n$. 

\subsection{Motivation}
Let $\hat{\theta}$ denote the unknown network parameters. In general, a strong adversarial example should  have high success rates on both white-box models,~\ie,~$L(X^{adv}, y^{\text{true}}; \theta) > L(X, y^{\text{true}}; \theta)$, and black-box models,~\ie,~$L(X^{adv}, y^{\text{true}}; \hat{\theta}) > L(X, y^{\text{true}}; \hat{\theta})$. On one hand, the traditional single-step attacks,~\eg,~FGSM, tend to underfit to the specific network parameters $\theta$ due to inaccurate linear appropriation of the loss $L(X, y^{\text{true}}; \theta)$, thus cannot reach high success rates on white-box models. On the other hand, the traditional iterative attacks,~\eg,~I-FGSM, greedily perturb the images in the direction of the sign of the loss gradient $\nabla_{X} L(X, y^{\text{true}}; \theta)$ at each iteration, and thus easily fall into the poor local maxima and overfit to the specific network parameters $\theta$. These overfitted adversarial examples rarely transfer to black-box models. In order to generate adversarial examples with strong transferability, we need to find a better way to optimize the loss $L(X, y^{\text{true}}; \theta)$ to alleviate this overfitting phenomenon.

Data augmentation~\cite{krizhevsky2012imagenet,simonyan2015very,he2016identity} is shown as an effective way to prevent networks from overfitting during the training process. Meanwhile, \cite{xie2017mitigating,guo2017countering} showed that adversarial examples are no longer malicious if simple image transformations are applied, which indicates these transformed adversarial images can serve as good samples for better optimization. Those facts inspire us to apply random and differentiable transformations to the inputs for the sake of the transferability of adversarial examples.

\subsection{Diverse Input Patterns}
Based on the analysis above, we aim at generating more transferable adversarial examples via diverse input patterns.

\paragraph{DI\textsuperscript{2}-FGSM.} First, we propose the Diverse Inputs Iterative Fast Gradient Sign Method (DI\textsuperscript{2}-FGSM), which applies image transformations $T(\cdot)$ to the inputs with the probability $p$ at each iteration of I-FGSM~\cite{kurakin2016scale} to alleviate the overfitting phenomenon. 

In this paper, we consider random resizing, which resizes the input images to a random size, and random padding, which pads zeros around the input images in a random manner~\cite{xie2017mitigating}, as the instantiation of the image transformations $T(\cdot)$\footnote{We have also experimented with other image transformations, \eg, rotation or flipping, to create diverse input patterns, and found random resizing \& padding yields adversarial examples with the \emph{best} transferability.}. The transformation probability $p$ controls the trade-off between success rates on white-box models and success rates on black-box models, which can be observed from Fig.~\ref{fig: transformation probability}. If $p=0$, DI\textsuperscript{2}-FGSM degrades to I-FGSM and leads to overfitting. If $p=1$,~\ie,~only transformed inputs are used for the attack, the generated adversarial examples tend to have much higher success rates on black-box models but lower success rates on white-box models, since the original inputs are not seen by the attackers. 

In general, the updating procedure of DI\textsuperscript{2}-FGSM is similar to I-FGSM, with the replacement of Eq.~\eqref{eq:IFGSM update} by
\begin{equation} \small
X_{n+1}^{\text{adv}}=\text{Clip}_{X}^{\epsilon}\{X_{n}^{adv}+\alpha\cdot\text{sign}\left(\nabla_{X} L(T(X_{n}^{adv};p),y^{\text{true}};\theta)\right)\},
\end{equation}
where the stochastic transformation function $T(X_{n}^{adv};p)$ is
\begin{equation}
T(X_{n}^{adv};p)=
\begin{cases}
T(X_{n}^{adv})~&\text{with probability $p$} \\ 
X_{n}^{adv}~&\text{with probability $1-p$} 
\end{cases}.
\end{equation}

\paragraph{M-DI\textsuperscript{2}-FGSM.}~Intuitively, momentum and diverse inputs are two completely different ways to alleviate the overfitting phenomenon. We can combine them naturally to form a much stronger attack,~\ie,~Momentum Diverse Inputs Iterative Fast Gradient Sign Method (M-DI\textsuperscript{2}-FGSM). The overall updating procedure of M-DI\textsuperscript{2}-FGSM is similar to MI-FGSM, with the only replacement of Eq.~\eqref{eq: momentum term} by
\begin{equation}
g_{n+1} = \mu \cdot g_n + \frac{\nabla_{X} L(T(X_{n}^{adv};p), y^{\text{true}}; \theta)}{||\nabla_{X} L(T(X_{n}^{adv};p), y^{\text{true}}; \theta)||_1}.
\end{equation}

\subsection{Relationships between Different Attacks}
\begin{figure}
\centering
\includegraphics[width=0.35\textwidth]{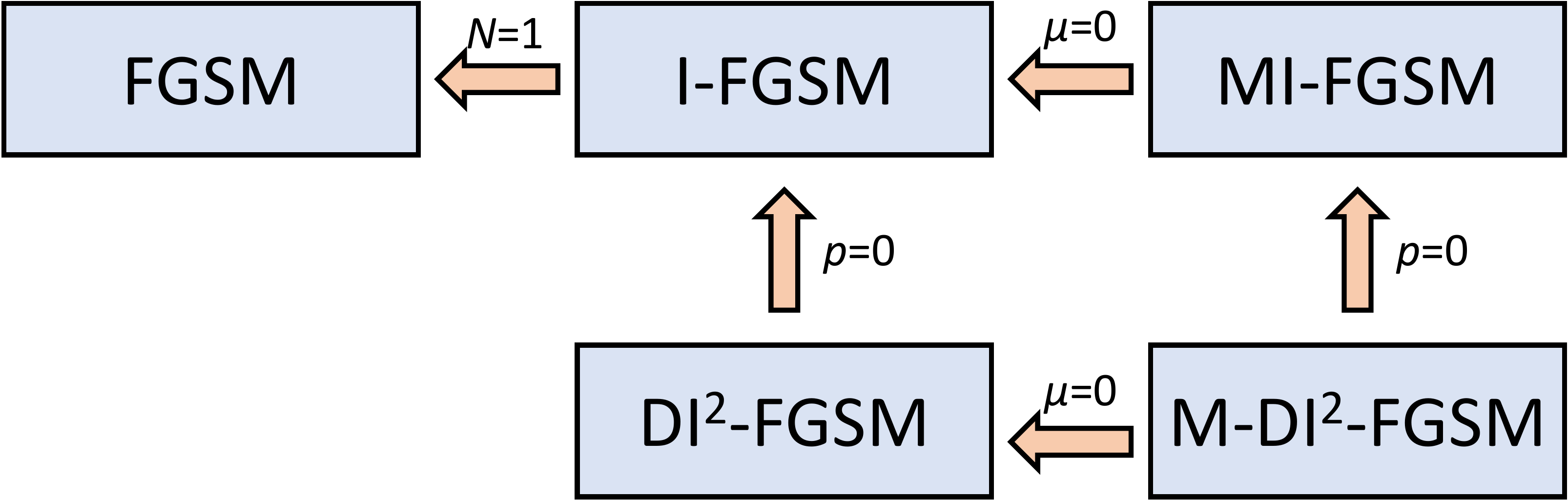}
\caption{\textbf{Relationships between different attacks}. By setting setting values of the transformation probability $p$, the decay factor $\mu$ and the total iteration number $N$, we can relate these different attacks in the family of Fast Gradient Sign Methods.}
\label{fig:relationship}
\vspace{-2ex}
\end{figure}

The attacks mentioned above all belong to the family of Fast Gradient Sign Methods, and they can be related via different parameter settings as shown in Fig.~\ref{fig:relationship}. To summarize, 
\begin{itemize}
\item If the transformation probability $p=0$, M-DI\textsuperscript{2}-FGSM degrades to MI-FGSM, and DI\textsuperscript{2}-FGSM degrades to I-FGSM.
\item If the decay factor $\mu=0$, M-DI\textsuperscript{2}-FGSM degrades to DI\textsuperscript{2}-FGSM, and MI-FGSM degrades to I-FGSM.
\item If the total iteration number $N=1$, I-FGSM degrades to FGSM.
\end{itemize}

\subsection{Attacking an Ensemble of Networks }
Liu~\etal~\cite{liu2016delving} suggested that attacking an ensemble of multiple networks simultaneously can generate much stronger adversarial examples. The motivation is that if an adversarial image remains adversarial for multiple networks, then it is more likely to transfer to other networks as well. Therefore, we can use this strategy to improve the transferability even further.

We follow the ensemble strategy proposed in \cite{dong2017boosting}, which fuse the logit activations together to attack multiple networks simultaneously. Specifically, to attack an ensemble of $K$ models, the logits are fused by:
\begin{equation}
l(X; \theta_1, ..., \theta_K) = \sum\limits_{k=1}^K w_k l_k(X; \theta_k)
\end{equation}
where $l_k(X; \theta_k)$ is the logits output of the $k$-th model with the parameters $\theta_k$, $w_k$ is the ensemble weight with $w_k \geq 0$ and $\sum\limits_{k=1}^K w_k= 1$.

\section{Experiment} \label{sec:exp}

\subsection{Experiment Setup}
\paragraph{Dataset.}~It is less meaningful to attack the images that are already classified wrongly. Therefore, we randomly choose $5000$ images from the ImageNet validation set that are classified correctly by all the networks which we test on, to form our test dataset. All these images are resized to $299 \times 299 \times 3$ beforehand.

\paragraph{Networks.}~We consider four normally trained networks,~\ie,~Inception-v3 (Inc-v3)~\cite{szegedy2016rethinking}, Inception-v4 (Inc-v4)~\cite{szegedy2017inception}, Resnet-v2-152 (Res-152)~\cite{he2016identity} and Inception-Resnet-v2 (IncRes-v2)~\cite{szegedy2017inception}, and three adversarially trained networks~\cite{tramer2017ensemble},~\ie,~ens3-adv-Inception-v3 (Inc-v3\textsubscript{ens3}), ens4-adv-Inception-v3 (Inc-v3\textsubscript{ens4}) and ens-adv-Inception-ResNet-v2  (IncRes-v2\textsubscript{ens}). All networks are publicly available\footnote{\url{https://github.com/tensorflow/models/tree/master/research/slim}}\textsuperscript{,}\footnote{\url{https://github.com/tensorflow/models/tree/master/research/adv_imagenet_models}}.

\paragraph{Implementation details.}~For the parameters of different attackers, we follow the default settings in \cite{kurakin2016adversarial} with the step size $\alpha = 1$ and the total iteration number $N = \min (\epsilon+4, 1.25\epsilon)$. We set the maximum perturbation of each pixel to be $\epsilon = 15$, \emph{which is still imperceptible for human observers}~\cite{lou2016foveation}. For the momentum term, decay factor $\mu$ is set to be $1$ as in \cite{dong2017boosting}. For the stochastic transformation function $T(X;p)$, the probability $p$ is set to be $0.5$,~\ie,~attackers put equal attentions on the original inputs and the transformed inputs. For transformation operations $T(\cdot)$, the input $X$ is first randomly resized to a $rnd \times rnd \times 3$ image, with $rnd \in [299, 330)$, and then padded to the size $ 330 \times 330 \times 3$ in a random manner.  

\subsection{Attacking a Single Network} \label{sec: single model}
\begin{table*}[tb]
\small
\centering
\begin{tabular}{|l|l|ccccccc|}
\hline
Model  & Attack    & Inc-v3 & Inc-v4 & IncRes-v2 & Res-152 & Inc-v3\textsubscript{ens3} & Inc-v3\textsubscript{ens4} & IncRes-v2\textsubscript{ens}  \\ \hline\hline
\multirow{5}{*}{Inc-v3}    & FGSM      & 64.6\% & 23.5\% & 21.7\%    & 21.7\%  & 8.0\%       & 7.5\%       & 3.6\%         \\
                           & I-FGSM    & \textbf{99.9\%} & 14.8\% & 11.6\%    & 8.9\%   & 3.3\%       & 2.9\%       & 1.5\%         \\ 
                           & DI\textsuperscript{2}-FGSM (\textbf{Ours})   & \textbf{99.9\%} & 35.5\% & 27.8\%    & 21.4\%  & 5.5\%       & 5.2\%       & 2.8\% \\
                           & MI-FGSM   & \textbf{99.9\%} & 36.6\% & 34.5\%    & 27.5\%  & 8.9\%       & 8.4\%       & 4.7\%         \\ 
                           & M-DI\textsuperscript{2}-FGSM (\textbf{Ours}) & \textbf{99.9\%} & \textbf{63.9\%} & \textbf{59.4\%} & \textbf{47.9\%}  & \textbf{14.3\%}  & \textbf{14.0\%}  & \textbf{7.0\%}    \\ \hline\hline
\multirow{5}{*}{Inc-v4}    & FGSM      & 26.4\% & 49.6\% & 19.7\%    & 20.4\%  & 8.4\%       & 7.7\%       & 4.1\%         \\
                           & I-FGSM    & 22.0\% & \textbf{99.9\%} & 13.2\%    & 10.9\%  & 3.2\%       & 3.0\%       & 1.7\%         \\ 
                           & DI\textsuperscript{2}-FGSM (\textbf{Ours})   & 43.3\% & 99.7\% & 28.9\%    & 23.1\%  & 5.9\%       & 5.5\%       & 3.2\%         \\ 
                           & MI-FGSM   & 51.1\% & \textbf{99.9\%} & 39.4\%    & 33.7\%  & 11.2\%      & 10.7\%      & 5.3\%         \\ 
                           & M-DI\textsuperscript{2}-FGSM (\textbf{Ours}) & \textbf{72.4\%} & 99.5\% & \textbf{62.2\%}    & \textbf{52.1\%}  & \textbf{17.6\%}      & \textbf{15.6\%}      & \textbf{8.8\%}        \\ \hline \hline
\multirow{5}{*}{IncRes-v2} & FGSM      & 24.3\% & 19.3\% & 39.6\%    & 19.4\%  & 8.5\%       & 7.3\%       & 4.8\%         \\ 
                           & I-FGSM    & 22.2\% & 17.7\% & 97.9\%    & 12.6\%  & 4.6\%       & 3.7\%       & 2.5\%         \\
                           & DI\textsuperscript{2}-FGSM (\textbf{Ours})   & 46.5\% & 40.5\% & 95.8\%    & 28.6\%  & 8.2\%       & 6.6\%       & 4.8\%         \\ 
                           & MI-FGSM   & 53.5\% & 45.9\% & \textbf{98.4\%}    & 37.8\%  & 15.3\%      & 13.0\%      & 8.8\%         \\ 
                           & M-DI\textsuperscript{2}-FGSM (\textbf{Ours}) & \textbf{71.2\%} & \textbf{67.4\%} & 96.1\%    & \textbf{57.4\%}  & \textbf{25.1\%} & \textbf{20.7\%}    & \textbf{14.9\%}        \\ \hline \hline
\multirow{5}{*}{Res-152}   & FGSM      & 34.4\% & 28.5\% & 27.1\%    & 75.2\%  & 12.4\%      & 11.0\%      & 6.0\%         \\ 
                           & I-FGSM    & 20.8\% & 17.2\% & 14.9\%    & 99.1\%  & 5.4\%       & 4.6\%       & 2.8\%         \\ 
                           & DI\textsuperscript{2}-FGSM (\textbf{Ours})   & 53.8\% & 49.0\% & 44.8\%    & \textbf{99.2\%}  & 13.0\%      & 11.1\%      & 6.9\% \\
                           & MI-FGSM   & 50.1\% & 44.1\% & 42.2\%    & 99.0\%  & 18.2\%      & 15.2\%      & 9.0\%         \\  
                           & M-DI\textsuperscript{2}-FGSM (\textbf{Ours}) & \textbf{78.9\%} & \textbf{76.5\%} & \textbf{74.8\%}    & \textbf{99.2\%}  & \textbf{35.2\%}   & \textbf{29.4\%}      & \textbf{19.0\%}  \\ \hline
\end{tabular}
\caption{\textbf{The success rates on seven networks where we attack a single network}. The diagonal blocks indicate white-box attacks, while the off-diagonal blocks indicate black-box attacks which are much more challenging. Experiment results demonstrate that our proposed input diversity strategy substantially improve the transferability of generated adversarial examples.}
\label{table:single model 15}
\vspace{-1ex}
\end{table*}

We first perform adversarial attacks on a single network. We craft adversarial examples only on normally trained networks, and test them on all seven networks. The success rates are shown in Table~\ref{table:single model 15}, where the diagonal blocks indicate white-box attacks and off-diagonal blocks indicate black-box attacks. We list the networks that we attack on in rows, and networks that we test on in columns. 

From Table~\ref{table:single model 15}, a first glance shows that M-DI\textsuperscript{2}-FGSM outperforms all other baseline attacks by a large margin on all black-box models, and maintains high success rates on all white-box models. For example, if adversarial examples are crafted on IncRes-v2, M-DI\textsuperscript{2}-FGSM has success rates of $67.4\%$ on Inc-v4 (normally trained black-box model) and $25.1\%$ on Inc-v3\textsubscript{ens3} (adversarially trained black-box model), while strong baselines like MI-FGSM only obtains the corresponding success rates of $45.9\%$ and $15.3\%$, respectively. This convincingly demonstrates the effectiveness of the combination of input diversity and momentum for improving the transferability of adversarial examples.

\begin{figure}[t!]
\centering
\includegraphics[width=\linewidth]{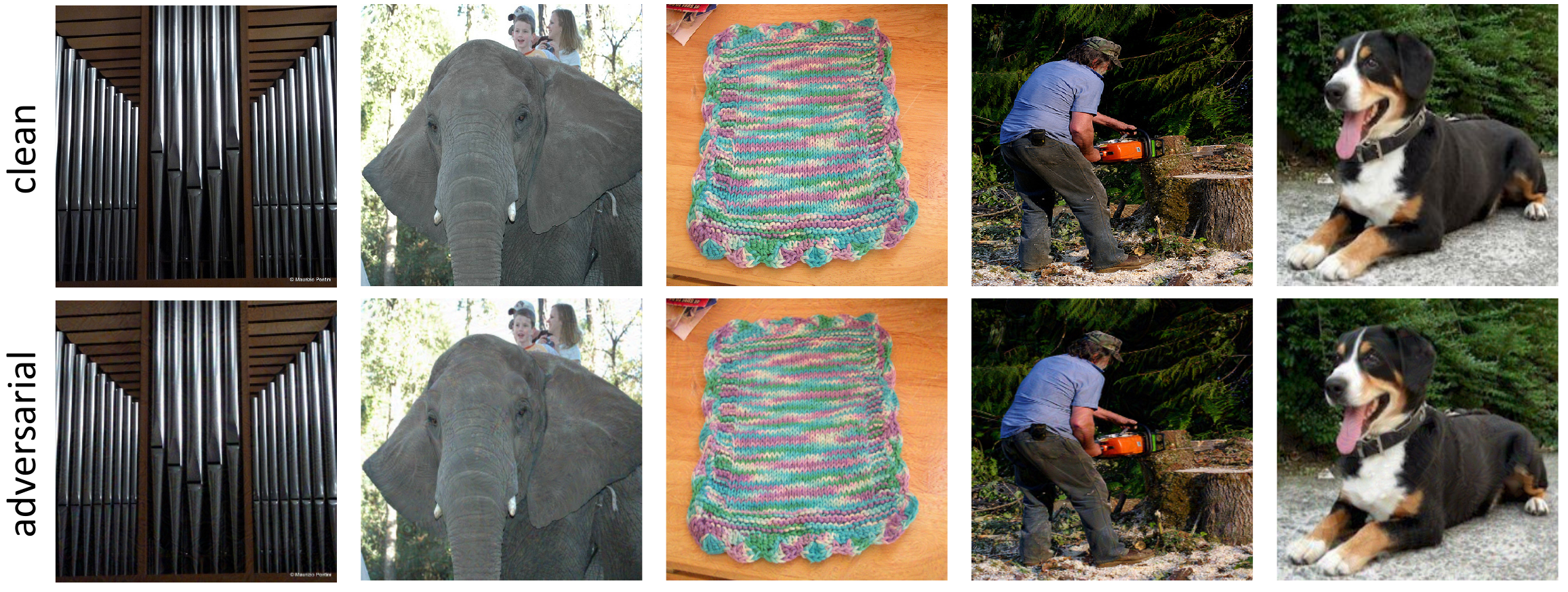}
\vspace{-1.5em}
\caption{\textbf{Visualization of randomly selected clean images and their corresponding adversarial examples}. All these adversarial examples are generated on Inception-v3 using our proposed DI\textsuperscript{2}-FGSM with the maximum perturbation of each pixel $\epsilon$ = 15.}
\label{fig:vis}
\vspace{-1em}
\end{figure}

We then compare the success rates of I-FGSM and DI\textsuperscript{2}-FGSM to see the effectiveness of diverse input patterns solely. By generating adversarial examples with input diversity, DI\textsuperscript{2}-FGSM significantly improves the success rates of I-FGSM on challenging black-box models, regardless whether this model is adversarially trained, and maintains high success rates on white-box models. For example, if adversarial examples are crafted on Res-152, DI\textsuperscript{2}-FGSM has success rates of $99.2\%$ on Res-152 (white-box model), $53.8\%$ on Inc-v3 (normally trained black-box model) and $11.1\%$ on Inc-v3\textsubscript{ens4} (adversarially trained black-box model), while I-FGSM only obtains the corresponding success rates of $99.1\%$, $20.8\%$ and $4.6\%$, respectively. Compared with FGSM, DI\textsuperscript{2}-FGSM also reaches much higher success rates on the normally trained black-box models, and comparable performance on the adversarially trained black-box models. Besides, we visualize 5 randomly selected pairs of such generated adversarial images and their clean counterparts in Figure \ref{fig:vis}. These visualization results show that these generated adversarial perturbations are human imperceptible.

\begin{table*}[h!]
\centering
\small
\begin{tabular}{|l|p{2.5cm}|ccccccc|}
\hline
Model  & Attack    & Inc-v3 & Inc-v4 & IncRes-v2 & Res-152 & Inc-v3\textsubscript{ens3} & Inc-v3\textsubscript{ens4} & IncRes-v2\textsubscript{ens} \\ \hline\hline
\multirow{2}{*}{Inc-v3}  &  C\&W  & \textbf{100.0\%} & 5.7\%   & 5.3\%   & 5.1\%   & 3.0\%  & 2.5\% & 1.1\%    \\ 
                         & D-C\&W (\textbf{Ours})  & \textbf{100.0\%} & \textbf{16.8\%}  & \textbf{13.0\%}  & \textbf{11.2\%}  & \textbf{5.8\%}  & \textbf{3.9\%} & \textbf{2.1\%}    \\
\hline
\multirow{2}{*}{Inc-v4}    
                           & C\&W  & 15.1\%  & \textbf{100.0\%} & 9.2\%   & 7.8\%   & 4.4\%  & 3.5\% & 1.9\%   \\ 
                           & D-C\&W (\textbf{Ours})   & \textbf{29.3\%}  & \textbf{100.0\%} & \textbf{20.1\%}  & \textbf{15.4\%}  & \textbf{7.1\%}  & \textbf{5.3\%} & \textbf{3.1\%}  \\ 
\hline
\multirow{2}{*}{IncRes-v2}  & C\&W   & 15.8\%  & 11.2\%  & 99.9\%  & 8.6\%   & 6.3\%  & 3.6\% & 3.4\%  \\ 
                           & D-C\&W (\textbf{Ours})  & \textbf{33.9\%}  & \textbf{25.6\%}  & \textbf{100.0\%} & \textbf{19.4\%}  & \textbf{11.2\%} & \textbf{7.3\%} & \textbf{4.0\%} \\ 
\hline
\multirow{2}{*}{Res-152}   
                           & C\&W   & 11.4\%  & 6.9\%   & 6.1\%   & \textbf{100.0\%} & 4.4\%  & 4.1\% & 2.3\%   \\ 
                           & D-C\&W (\textbf{Ours})   & \textbf{33.0\%}  & \textbf{27.7\%}  & \textbf{24.4\%}  & \textbf{100.0\%} & \textbf{13.1\%} & \textbf{9.3\%} & \textbf{5.7\%}  \\ 
\hline
\end{tabular}
\caption{\textbf{The success rates on seven networks where we attack a single network using C\&W attack}. Experiment results demonstrate that the proposed input diversity strategy can enhance C\&W attack for generating more transferable adversarial examples.}
\label{table:single model cw}
\vspace{-1ex}
\end{table*}

It should be mentioned that the proposed input diversity is not merely applicable to fast gradient sign methods. To demonstrate the generalization, we also incorporate C\&W attack~\cite{carlini2016towards} with input diversity. The experiment is conducted on $1000$ correctly classified images. For the parameters of C\&W, the maximal iteration is $250$, the learning rate is $0.01$ and the confidence is $10$. As Table~\ref{table:single model cw} suggests, our method D-C\&W obtains a significant performance improvement over C\&W on black-box models.

\subsection{Attacking an Ensemble of Networks}
\begin{table*}[tb]
\small
\centering
\begin{tabular}{|l|l|ccccccc|}
\hline
Model & Attack    & -Inc-v3 & -Inc-v4 & -IncRes-v2 & -Res-152 & -Inc-v3\textsubscript{ens3} & -Inc-v3\textsubscript{ens4} & -IncRes-v2\textsubscript{ens} \\ \hline\hline
\multirow{4}{*}{Ensemble} & I-FGSM    & 96.6\%  & \textbf{96.9\%}  & 98.7\%     & 96.2\%   & \textbf{97.0\%}       & \textbf{97.3\%}       & 94.3\%         \\ 
                          & DI\textsuperscript{2}-FGSM (\textbf{Ours})   & 88.9\%  & 89.6\%  & 93.2\%     & 87.7\%   & 91.7\%       & 91.7\%       & 93.2\%         \\ 
                          & MI-FGSM   & \textbf{96.9\%}  & \textbf{96.9\%}  &\textbf{98.8\%}     & \textbf{96.8\%}   & 96.8\%       & 97.0\%       & 94.6\%         \\ 
                          & M-DI\textsuperscript{2}-FGSM (\textbf{Ours}) & 90.1\%  & 91.1\%  & 94.0\%     & 89.3\%   & 92.8\%       & 92.7\%       & \textbf{94.9\%}         \\ \hline \hline
\multirow{4}{*}{Hold-out} & I-FGSM    & 43.7\%  & 36.4\%  & 33.3\%     & 25.4\%   & 12.9\%       & 15.1\%       & 8.8\%          \\ 
                          & DI\textsuperscript{2}-FGSM (\textbf{Ours})   & 69.9\%  & 67.9\%  & 64.1\%     & 51.7\%   & 36.3\%       & 35.0\%       & 30.4\%         \\ 
                          & MI-FGSM   & 71.4\%  & 65.9\%  & 64.6\%     & 55.6\%   & 22.8\%       & 26.1\%       & 15.8\%         \\ 
                          & M-DI\textsuperscript{2}-FGSM (\textbf{Ours}) & \textbf{80.7\%}  & \textbf{80.6\%}  & \textbf{80.7\%}     & \textbf{70.9\%}   & \textbf{44.6\%}   & \textbf{44.5\%} & \textbf{39.4\%}         \\ \hline
\end{tabular}
\caption{\textbf{The success rates of ensemble attacks}. Adversarial examples are generated on an ensemble of six networks, and tested on the ensembled network (\textit{white-box setting}) and the hold-out network (\textit{black-box setting}). The sign ``-'' indicates the hold-out network. We observe that the proposed M-DI\textsuperscript{2}-FGSM significantly outperforms \emph{all} other attacks on \emph{all} black-box models.}
\label{table: ensemble model 15}
\vspace{-2.7ex}
\end{table*}

Though the results in Table~\ref{table:single model 15} show that momentum and input diversity can significantly improve the transferability of adversarial examples, they are still relatively weak at attacking an adversarially trained network under the black-box setting,~\eg,~the highest black-box success rate on IncRes-v2\textsubscript{ens} is only $19.0\%$. Therefore, we follow the strategy in \cite{liu2016delving} to attack multiple networks simultaneously in order to further improve transferability. We consider all seven networks here. Adversarial examples are generated on an ensemble of six networks, and tested on the ensembled network and the hold-out network, using I-FGSM, DI\textsuperscript{2}-FGSM, MI-FGSM and  M-DI\textsuperscript{2}-FGSM, respectively. FGSM is ignored here due to its low success rates on white-box models. All ensembled models are assigned with equal weight,~\ie,~$w_k = 1/6$.

The results are summarized in Table~\ref{table: ensemble model 15}, where the top row shows the success rates on the ensembled network (white-box setting), and the bottom row shows the success rates on the hold-out network (black-box setting).  Under the challenging black-box setting, we observe that M-DI\textsuperscript{2}-FGSM always generates adversarial examples with better transferability than other methods on all networks. For example, by keeping Inc-v3\textsubscript{ens3} as a hold-out model, M-DI\textsuperscript{2}-FGSM can fool Inc-v3\textsubscript{ens3} with an success rate of $44.6\%$, while I-FGSM, DI\textsuperscript{2}-FGSM and MI-FGSM only have success rates of $12.9\%$, $36.3\%$ and $22.8\%$, respectively. Besides, compared with MI-FGSM, we observe that using diverse input patterns alone,~\ie,~DI\textsuperscript{2}-FGSM, can reach a much higher success rate if the hold-out model is an adversarially trained network, and a comparable success rate if the hold-out model is a normally trained network.

Under the white-box setting, we see that DI\textsuperscript{2}-FGSM and M-DI\textsuperscript{2}-FGSM reach slightly lower (but still very high) success rates on ensemble models compared with I-FGSM and MI-FGSM. This is due to the fact that attacking multiple networks simultaneously is much harder than attacking a single model. However, the white-box success rates can be improved if we assign the transformation probability $p$ with a smaller value, increase the number of total iteration $N$ or use a smaller step size $\alpha$ (see Sec.~\ref{sec: ablation}). 

\subsection{Ablation Studies} \label{sec: ablation}
In this section, we conduct a series of ablation experiments to study the impact of different parameters. We only consider attacking an ensemble of networks here, since it is much stronger than attacking a single network and can provide a more accurate evaluation of the network robustness. The max perturbation of each pixel $\epsilon$ is set to $15$ for all experiments.

\paragraph{Transformation probability $p$.}~We first study the influence of the transformation probability $p$ on the success rates under both white-box and black-box settings. We set the step size $\alpha=1$ and the total iteration number $N=\min(\epsilon+4, 1.25\epsilon)$. The transformation probability $p$ varies from $0$ to $1$. Recall the relationships shown in Fig.~\ref{fig:relationship}, M-DI\textsuperscript{2}-FGSM (or DI\textsuperscript{2}-FGSM) degrades to MI-FGSM (or I-FGSM) if $p=0$.
\begin{figure}[tb]
\centering
\subfigure[]
{
\begin{minipage}[tb]{0.225\textwidth}
\includegraphics[width = 1\textwidth]{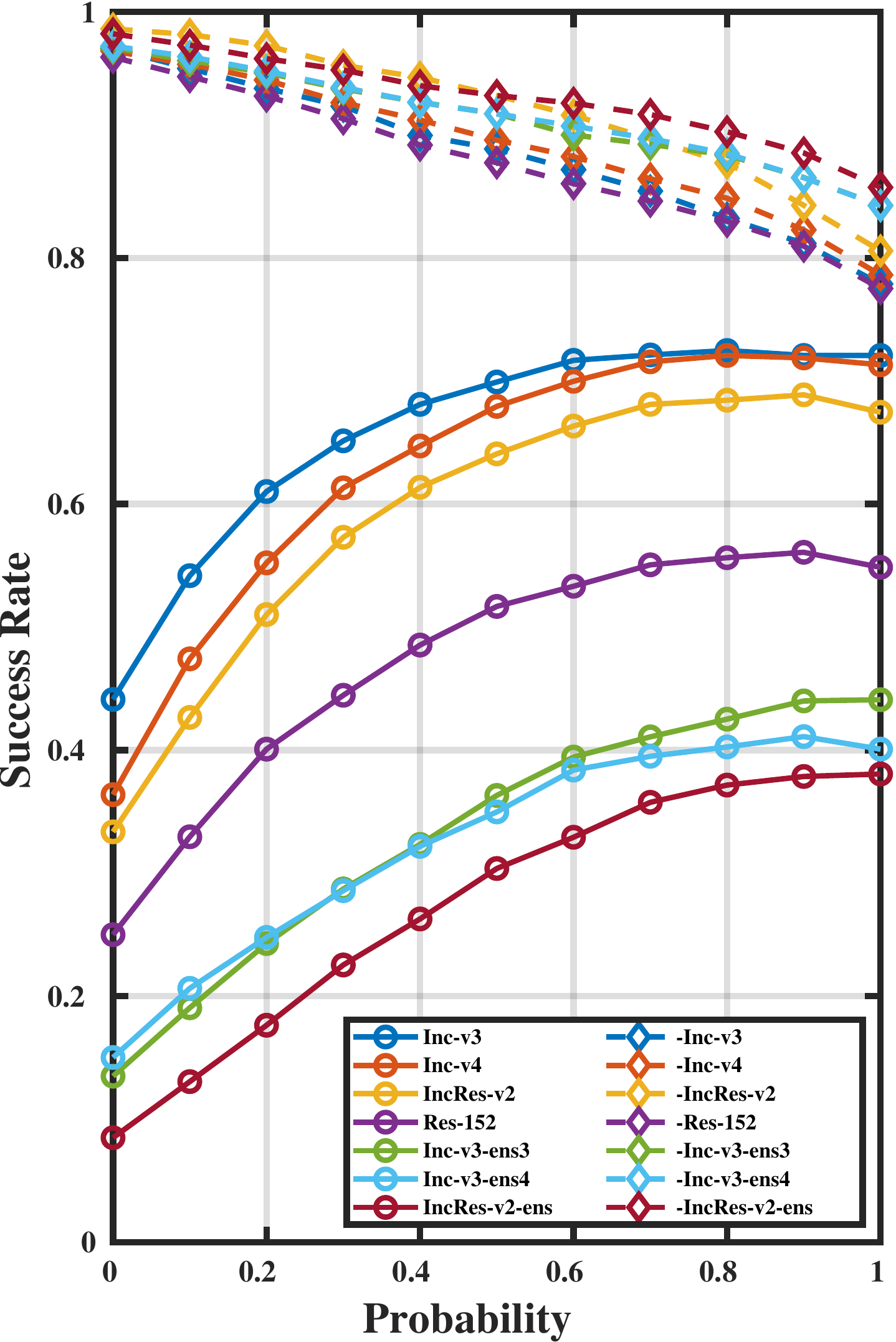}
\end{minipage}
}
\subfigure[]
{
\begin{minipage}[tb]{0.225\textwidth}
\includegraphics[width = 1\textwidth]{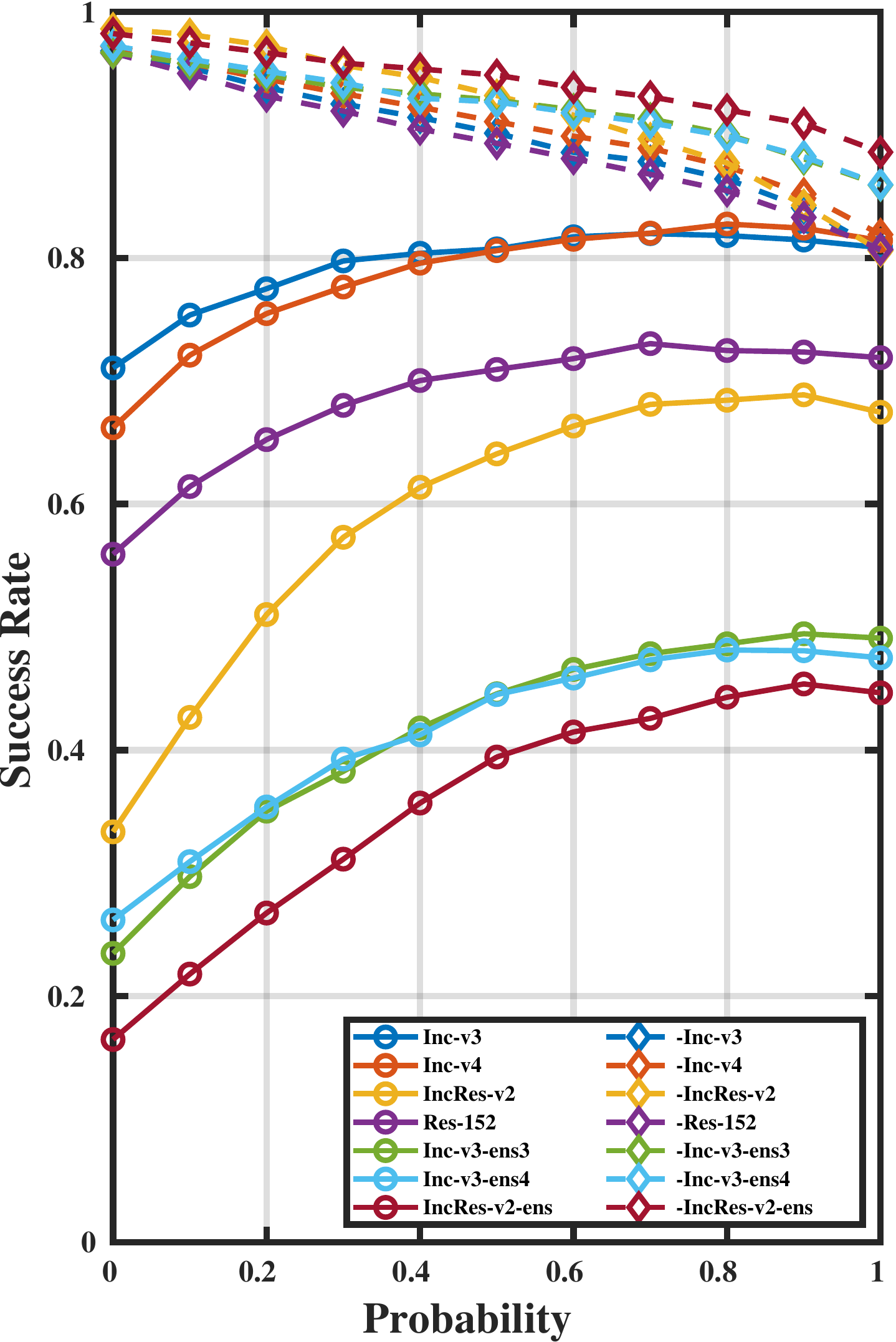}
\end{minipage}
}
\vspace{-2ex}
\caption{\textbf{The success rates of DI\textsuperscript{2}-FGSM (a) and M-DI\textsuperscript{2}-FGSM (b) when varying the transformation probability $p$}. ``Ensemble" (white-box setting) is with dashed lines and ``Hold-out" (black-box setting) is with solid lines.} 
\label{fig: transformation probability}
\vspace{-2ex}
\end{figure}

We show the success rates on various networks in Fig.~\ref{fig: transformation probability}. We observe that both DI\textsuperscript{2}-FGSM and M-DI\textsuperscript{2}-FGSM achieve a higher black-box success rates but lower white-box success rates as $p$ increase. Moreover, for all attacks, if $p$ is small,~\ie,~only a small amount of transformed inputs are utilized, black-box success rates can increase significantly, while white-box success rates only drop a little. This phenomenon reveals the importance of adding transformed inputs into the attack process.

\begin{figure}[tb]
\centering
\subfigure[]
{
\begin{minipage}[tb]{0.225\textwidth}
\includegraphics[width = 1\textwidth]{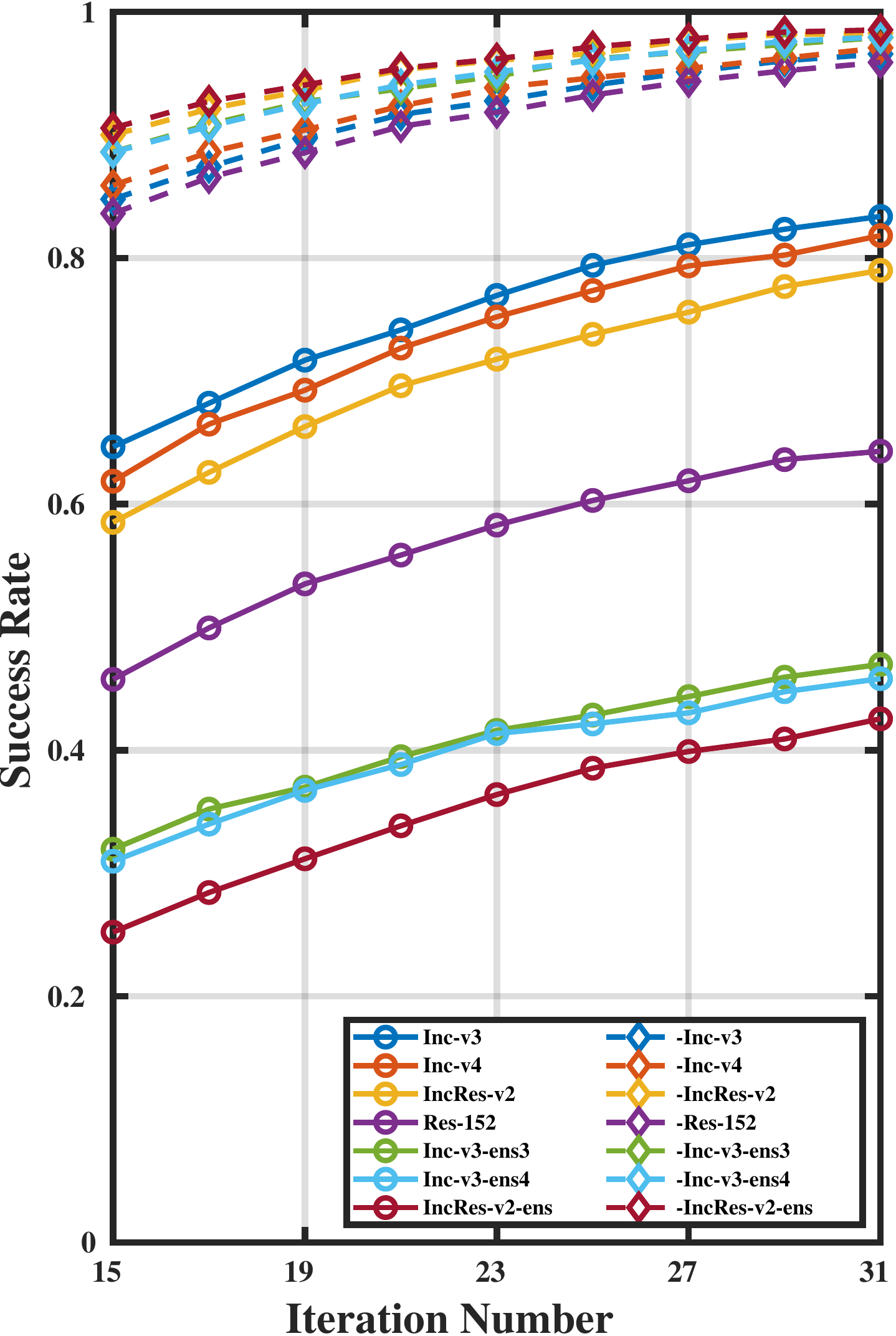}
\end{minipage}
}
\subfigure[]
{
\begin{minipage}[tb]{0.225\textwidth}
\includegraphics[width = 1\textwidth]{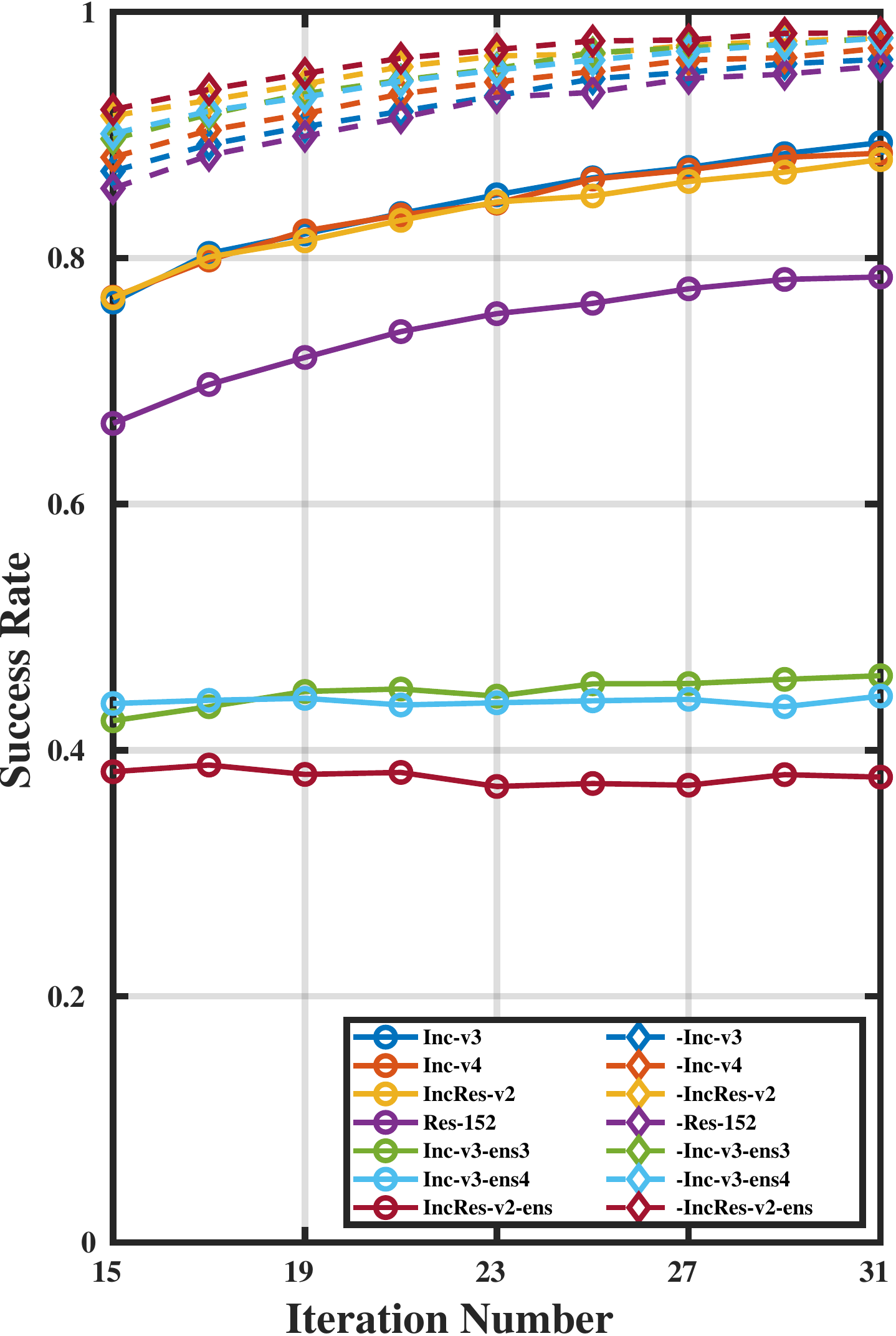}
\end{minipage}
}
\vspace{-2ex}
\caption{\textbf{The success rates of DI\textsuperscript{2}-FGSM (a) and M-DI\textsuperscript{2}-FGSM (b) when varying the total iteration number $N$}. ``Ensemble" (white-box setting) is with dashed lines and ``Hold-out" (black-box setting) is with solid lines.} 
\label{fig: iteration number} 
\vspace{-3ex}
\end{figure}

The trends shown in Fig.~\ref{fig: transformation probability} also provide useful suggestions of constructing strong adversarial attacks in practice. For example, if you know the black-box model is a new network that totally different from any existing networks, you can set $p=1$ to reach the maximum transferability. If the black-box model is a mixture of new networks and existing networks, you can choose a moderate value of $p$ to maximize the black-box success rates under a pre-defined white-box success rates,~\eg,~white-box success rates must greater or equal than $90\%$.

\paragraph{Total iteration number $N$.}~We then study the influence of the total iteration number $N$ on the success rates under both white-box and black-box settings. We set the transformation probability $p=0.5$ and the step size $\alpha=1$. The total iteration number $N$ varies from $15$ to $31$, and the results are plotted in Fig.~\ref{fig: iteration number}. For DI\textsuperscript{2}-FGSM, we see that the black-box success rates and white-box success rates always increase as the total iteration number $N$ increase. Similar trends can also be observed for M-DI\textsuperscript{2}-FGSM except for the black-box success rates on adversarially trained models,~\ie,~performing more iterations cannot bring extra transferability on adversarially trained models. Moreover, we observe that the success rates gap between M-DI\textsuperscript{2}-FGSM and DI\textsuperscript{2}-FGSM is diminished as $N$ increases.

\begin{figure}[tb]
\centering
\subfigure[]
{
\begin{minipage}[tb]{0.225\textwidth}
\includegraphics[width = 1\textwidth]{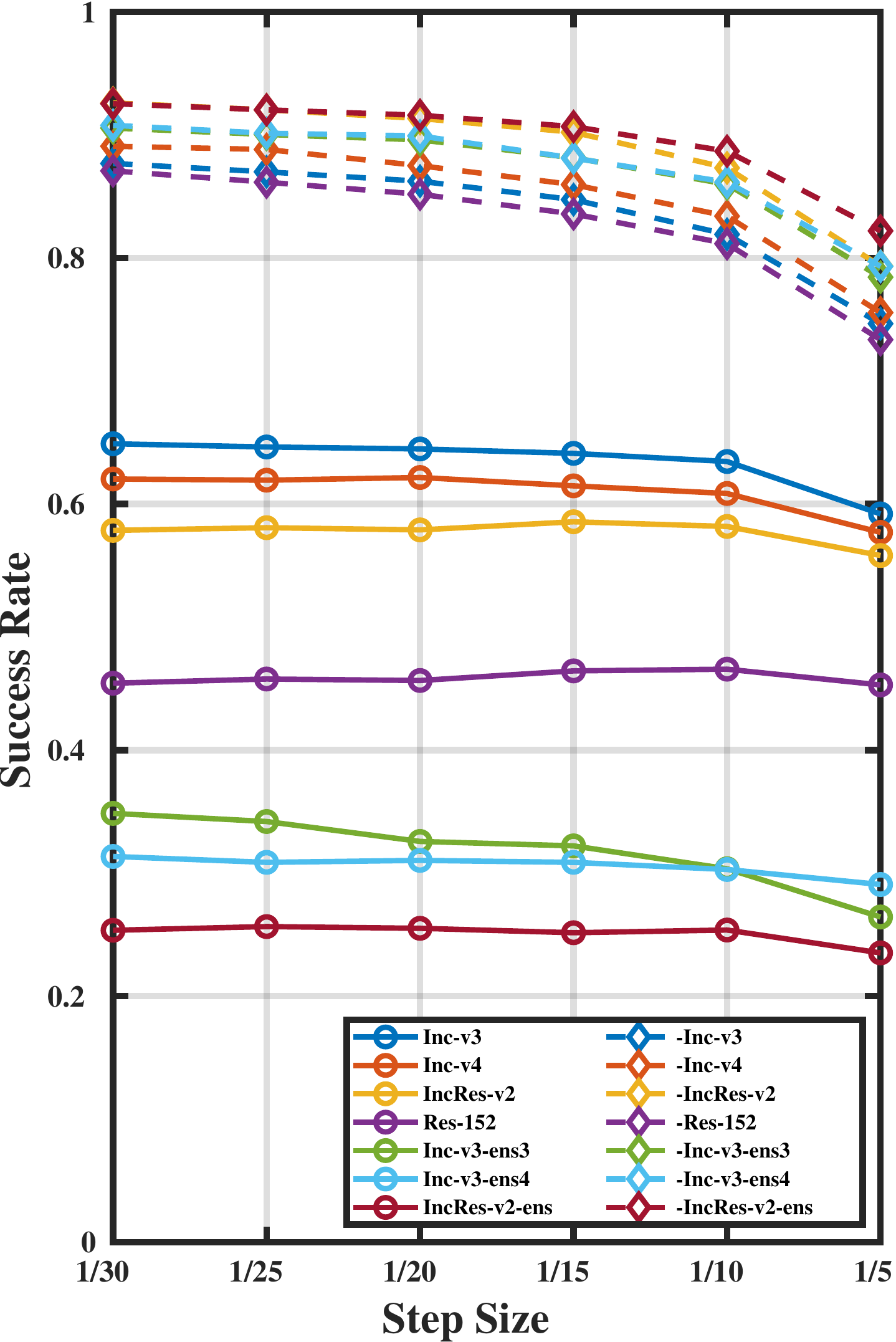}
\end{minipage}
}
\subfigure[]
{
\begin{minipage}[tb]{0.225\textwidth}
\includegraphics[width = 1\textwidth]{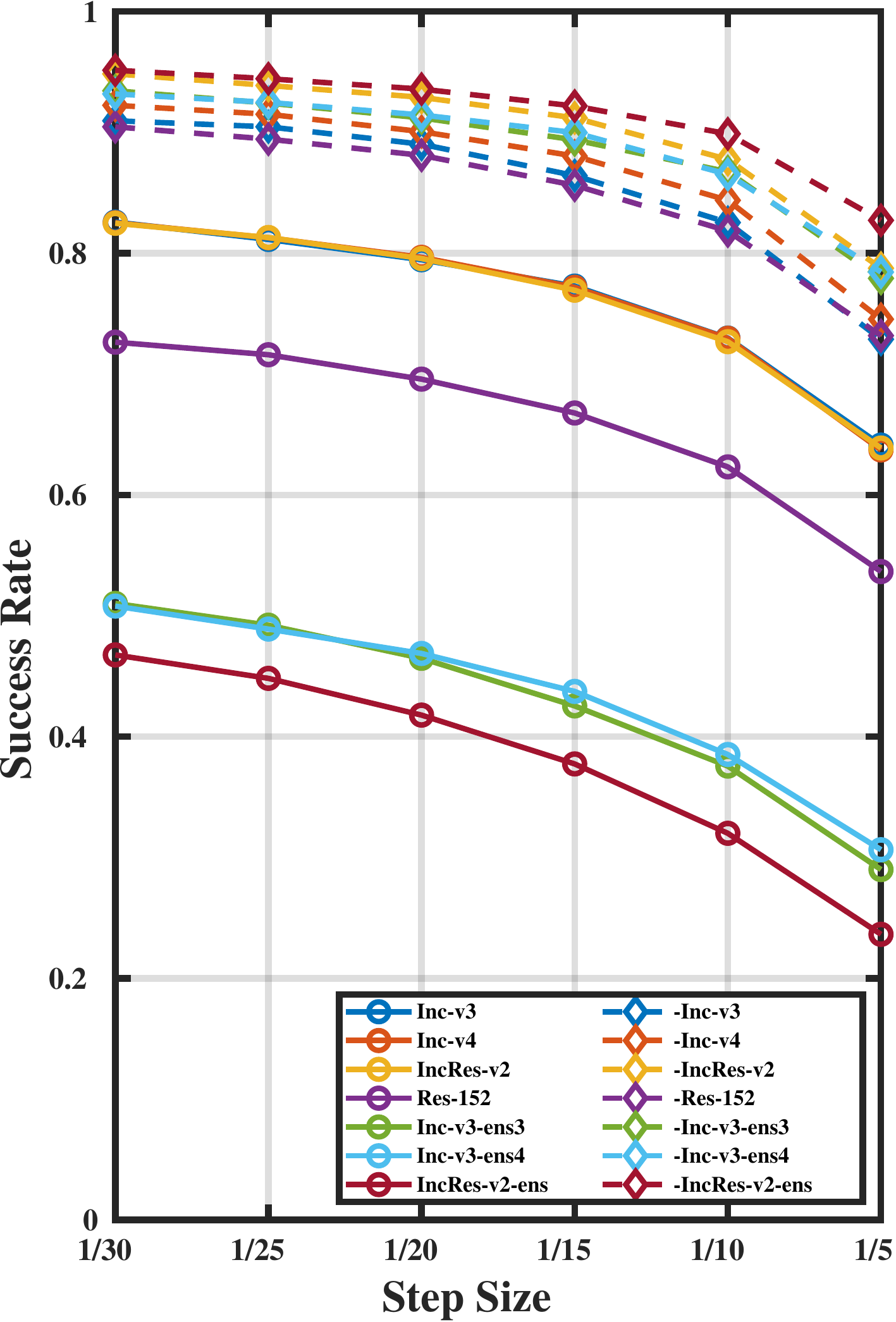}
\end{minipage}
}
\vspace{-2ex}
\caption{\textbf{The success rates of DI\textsuperscript{2}-FGSM (a) and M-DI\textsuperscript{2}-FGSM (b) when varying the step size $\alpha$}. ``Ensemble" (white-box setting) is with dashed lines and ``Hold-out" (black-box setting) is with solid lines.} 
\label{fig: step size}
\vspace{-3ex}
\end{figure}

\paragraph{Step size $\alpha$.}~We finally study the influence of the step size $\alpha$ on the success rates under both white-box and black-box settings. We set the transformation probability $p=0.5$. In order to reach the maximum perturbation $\epsilon$ even for a small step size $\alpha$, we set the total iteration number be proportional to the step size,~\ie,~$N=\epsilon/\alpha$. The results are plotted in Fig.~\ref{fig: step size}. We observe that the white-box success rates of both DI\textsuperscript{2}-FGSM and M-DI\textsuperscript{2}-FGSM can be boosted if a smaller step size is provided. Under the black-box setting, the success rates of DI\textsuperscript{2}-FGSM is insensitive to the step size, while the success rates of M-DI\textsuperscript{2}-FGSM can still be improved with smaller step size. 

\subsection{NIPS 2017 Adversarial Competition}
\begin{table*}[tb]
\small
\centering
\begin{tabular}{|l|ccccccc|}
\hline
Attack    & TsAIL  & iyswim & Anil Thomas & Inc-v3\textsubscript{adv} & IncRes-v2\textsubscript{ens} & Inc-v3 & \textbf{Average}\\ \hline\hline
I-FGSM    & 14.0\% & 35.6\% & 30.9\%      & \textbf{98.2\%}     & \textbf{96.4\%}        & \textbf{99.0\%}  & 62.4\% \\ %\hline
DI\textsuperscript{2}-FGSM (\textbf{Ours})  & \textbf{22.7\%} & 58.4\% & 48.0\%      & 91.5\%     & 90.7\%        & 97.3\%  & 68.1\%\\ %\hline
MI-FGSM   & 14.9\% & 45.7\% & 46.6\%      & 97.3\%     & 95.4\%        & 98.7\%  & 66.4\%\\ %\hline
MI-FGSM*   & 13.6\% & 43.2\% & 43.9\%      & 94.4\%     & 93.0\%        & 97.3\%  & 64.2\%\\ %\hline
M-DI\textsuperscript{2}-FGSM (\textbf{Ours}) & 20.0\% & \textbf{69.8\%} & \textbf{64.4\%}      & 93.3\%     & 92.4\%        & 97.9\%  & \textbf{73.0\%}\\ \hline
\end{tabular}
\caption{\textbf{The success rates on top defense solutions and official baselines from NIPS 2017 adversarial competition \cite{kurakin2018adversarial}}. * indicates the official results reported in the competition. Our proposed M-DI\textsuperscript{2}-FGSM reaches an average success rate of $73.0\%$, which outperforms the top-$1$ attack submission in the NIPS competition by a large margin of $6.6\%$.}
\vspace{-1.5em}
\label{table:nips}
\end{table*}

In order to verify the effectiveness of our proposed attack methods in practice, we here reproduce the top defense entries and official baselines from NIPS $2017$ adversarial competition~\cite{kurakin2018adversarial} for testing transferability. Due to the resource limitation, we only consider the top-$3$ defense entries,~\ie,~\emph{TsAIL}
% \footnote{\url{https://github.com/lfz/Guided-Denoise}}    
\cite{liao2018defense},  \emph{iyswim}
% \footnote{\url{https://github.com/cihangxie/NIPS2017\_adv\_challenge\_defense}} 
\cite{xie2017mitigating} and \emph{Anil Thomas}\footnote{\url{https://github.com/anlthms/nips-2017/tree/master/mmd}}, as well $3$ official baselines,~\ie,~Inc-v3\textsubscript{adv}, IncRes-v2\textsubscript{ens} and Inc-v3. We note that the No.1 solution and the No.3 solution apply significantly different image transformations (compared to random resizing \& padding used in our attack method) for defending against adversarial examples. For example, the No.1 solution, \emph{TsAIL}, applies an image denoising network for removing adversarial perturbations, and the No.3 solution, \emph{Anil Thomas}, includes a series of image transformations, \eg, JPEG compression, rotation, shifting and zooming, in the defense pipeline.
The test dataset contains $5000$ images which are all of the size $299\times299\times3$, and their corresponding labels are the same as the ImageNet labels.   

\paragraph{Generating adversarial examples.}~When generating adversarial examples, we follow the procedure in~\cite{kurakin2018adversarial}: (1) split the dataset equally into $50$ batches; (2) for each batch, the maximum perturbation $\epsilon$ is randomly chosen from the set $\{\frac{4}{255}, \frac{8}{255}, \frac{12}{255}, \frac{16}{255}\}$; and (3) generate adversarial examples for each batch under the corresponding $\epsilon$ constraint.

\paragraph{Attacker settings.}~For the settings of attackers, we follow \cite{dong2017boosting} by attacking an ensemble eight diferent models, \ie, Inc-v3, Inc-v4, IncRes-v2, Res-152, Inc-v3\textsubscript{ens3}, Inc-v3\textsubscript{ens4}, IncRes-v2\textsubscript{ens} and Inc-v3\textsubscript{adv}~\cite{kurakin2016scale}. The ensemble weights
are set as $1/7.25$ equally for the first seven models and $0.25/7.25$ for Inc-v3\textsubscript{adv}. The total iteration number $N$ is $10$ and the decay factor $\mu$ is $1$. This configuration for MI-FGSM won the $1$-st place in the NIPS $2017$ adversarial attack competition. For DI\textsuperscript{2}-FGSM and M-DI\textsuperscript{2}-FGSM, we choose $p=0.4$ according to the trends shown in Fig.~\ref{fig: transformation probability}.

\paragraph{Results.}~The results are summarized in Table~\ref{table:nips}. We also report the official results of MI-FGSM (named MI-FGSM*) as a reference to validate our implementation. The performance difference between MI-FGSM and MI-FGSM* is due to the randomness of the max perturbation magnitude introduced in the attack process. Compared with MI-FGSM, DI\textsuperscript{2}-FGSM have higher success rates on top defense solutions while slightly lower success rates on baseline models, which results in these two attack methods having similar average success rates. By integrating both diverse inputs and momentum term, this enhanced attack, M-DI\textsuperscript{2}-FGSM, reaches an average success rate of $73.0\%$, which is far better than other methods. For example, the top-$1$ attack submission, MI-FGSM, in the NIPS competition only gets an average success rate of $66.4\%$. We believe this superior transferability can also be observed on other defense submissions which we do not evaluate on.

\subsection{Discussion}
We provide a brief discussion of why the proposed diverse input patterns can help to generate adversarial examples with better transferability. One hypothesis is that the decision boundaries of different networks share similar inherent structures due to the same training dataset,~\eg, ImageNet. For example, as shown in Fig~\ref{fig:illustration}, different networks make similar mistakes in the presence of adversarial examples. By incorporating diverse patterns at each attack iteration, the optimization produces adversarial examples that are more robust to small transformations. These adversarial examples are malicious in a certain region at the network decision boundary, thus increasing the chance to fool other networks,~\ie,~they achieve better black-box success rate than existing methods. In the future, we plan to validate this hypothesis theoretically or empirically.

\section{Conclusions}
In this paper, we propose to improve the transferability of adversarial examples with input diversity. Specifically, our method applies random transformations to the input images at each iteration in the attack process. Compared with traditional iterative attacks, the results on ImageNet show that our proposed attack method gets significantly higher success rates for black-box models, and maintains similar success rates for white-box models. We improve the transferability further by integrating momentum term and attacking multiple networks simultaneously. By evaluating this enhanced attack against the top defense submissions and official baselines from NIPS $2017$ adversarial competition~\cite{kurakin2018adversarial}, we show that this enhanced attack reaches an average success rate of $73.0\%$, which outperforms the top-$1$ attack submission in the NIPS competition by a large margin of $6.6\%$. We hope that our proposed attack strategy can serve as a benchmark for evaluating the robustness of networks to adversaries and the effectiveness of different defense methods in future. Code is publicly available at \url{https://github.com/cihangxie/DI-2-FGSM}.

{\footnotesize
{\noindent {\bf Acknowledgement}: This work was supported by a gift grant from YiTu and ONR N00014-12-1-0883.}
}

{\small
\bibliographystyle{ieee}
\bibliography{egbib}
}

\end{document}